\documentclass[11pt]{article}
\usepackage{algorithm}
\usepackage{algpseudocode}
\usepackage[table]{xcolor}
\definecolor{cattlerow}{HTML}{F2F2F2}%
\definecolor{catline}{HTML}{C0C0C0}
\definecolor{qwenbg}{HTML}{F4F0FC}
\definecolor{llamabg}{HTML}{E6F2FF}
\usepackage[final]{acl}
\usepackage{amssymb}
\usepackage{times}
\usepackage{latexsym}
\usepackage{multirow}%
\usepackage{booktabs}%
\usepackage{graphicx}%
\newcommand{\qwenlogo}{%
    \raisebox{-0.15em}{\includegraphics[height=1.1em]{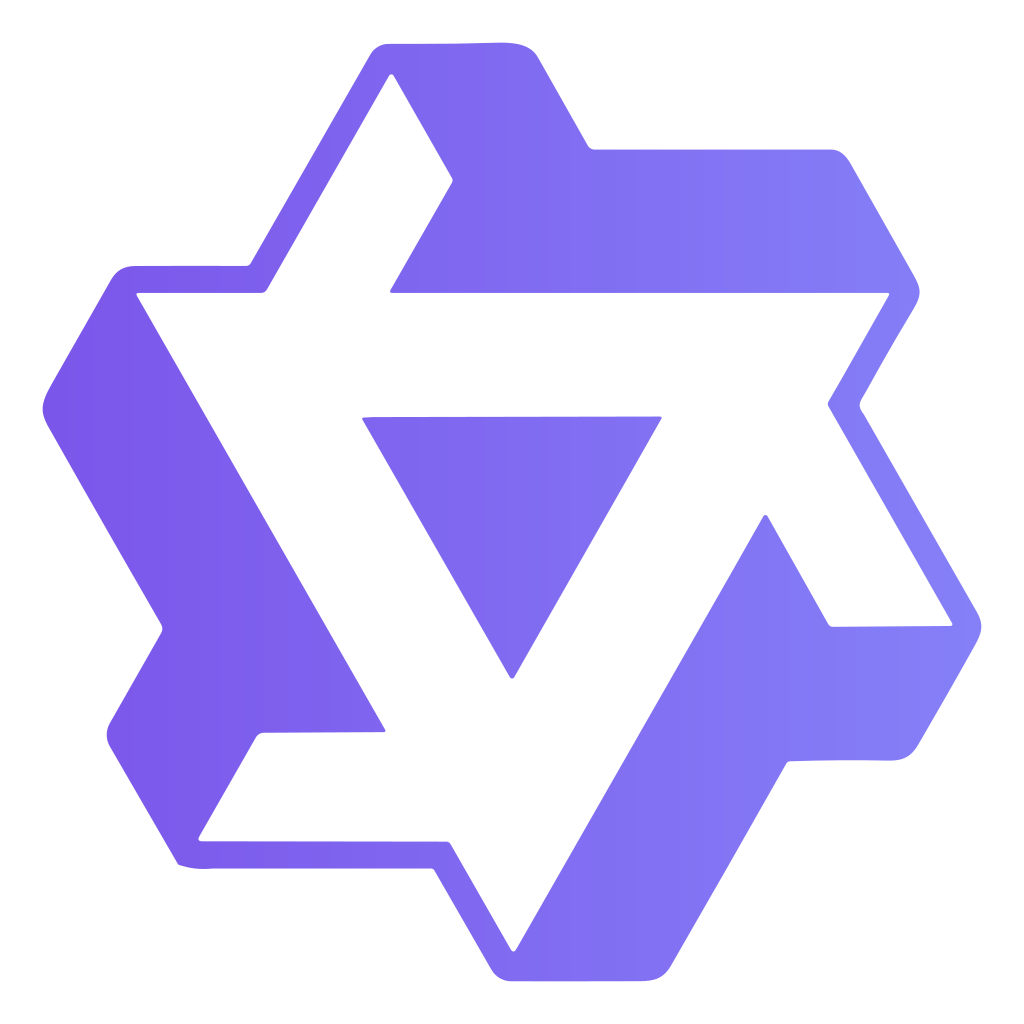}}%
    \hspace{0.3em}%
}
\newcommand{\llamalogo}{%
    \raisebox{-0.15em}{\includegraphics[height=1.1em]{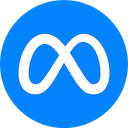}}%
    \hspace{0.3em}%
}
\usepackage[T1]{fontenc}
\usepackage{amsmath}%
\usepackage[utf8]{inputenc}

\usepackage{microtype}

\usepackage{inconsolata}

\usepackage{subcaption}

\title{FlashMem: Distilling Intrinsic Latent Memory via Computation Reuse}

\author{
    Yubo Hou\textsuperscript{1},
    Zhisheng Chen\textsuperscript{2,3},
    Tao Wan\textsuperscript{4},
    Zengchang Qin\textsuperscript{1,5}\thanks{Corresponding author.} \\
    \textsuperscript{1}School of ASEE, Beihang University, Beijing, China \\
    \textsuperscript{2}Institute of Computing Technology, Chinese Academy of Sciences, Beijing, China \\
    \textsuperscript{3}University of Chinese Academy of Sciences, Beijing, China \\
    \textsuperscript{4}School of BME, Beijing Advanced Innovation Center for BME, Beihang University \\
    \textsuperscript{5}CAIR and CECS, VinUniversity, Hanoi, Vietnam \\
    \texttt{houyubo@buaa.edu.cn, \textsuperscript{*}zengchang.qin@gmail.com}
}

\begin{document}
\maketitle

\begin{abstract} The stateless architecture of Large Language Models inherently lacks the mechanism to preserve dynamic context, compelling agents to redundantly reprocess history to maintain long-horizon autonomy. While latent memory offers a solution, current approaches are hindered by architectural segregation, relying on auxiliary encoders that decouple memory from the reasoning backbone. We propose \textbf{FlashMem}, a framework that distills intrinsic memory directly from transient reasoning states via computation reuse. Motivated by prior findings that final-layer representations preserve rich information about the input trajectory, FlashMem uses the last hidden state as a compact and empirically effective summary of the interaction history. This enables a \textbf{Shared-KV Consolidator} to synthesize memory by attending directly to the backbone's frozen cache, eliminating redundant re-parameterization. Furthermore, a parameter-free \textbf{Cognitive Monitor} leverages attention entropy to adaptively trigger consolidation only when high epistemic uncertainty is detected. Experiments demonstrate that FlashMem matches the performance of heavy baselines while reducing inference latency by \textbf{5 times}, effectively bridging the gap between efficiency and persistent cognition.\end{abstract}

\section{Introduction}
\label{sec:intro}

The capability to sustain coherent epistemic states across extended temporal horizons is a prerequisite for general-purpose autonomy.
While Large Language Models (LLMs) function as powerful reasoning engines, their architecture is fundamentally stateless: they map inputs to outputs based on static parameters without retaining a persistent internal state between interactions.
This design creates a dichotomy between the model's frozen parametric knowledge and the dynamic procedural context required for long-horizon tasks.
Consequently, in evolving workflows, agents are forced to discard rich intermediate reasoning representations after every inference step~\cite{hu2025memory}.
This statelessness necessitates the redundant reprocessing of historical information, creating a significant bottleneck for agents that must continually accumulate experience and adapt.

To eliminate the gap between transient inference and persistent cognition, memory mechanisms have become a focal point of research.
Current approaches are generally categorized into parametric, token-level, and latent forms~\cite{hu2025memory}.
Parametric updates are computationally prohibitive for real-time adaptation, while token-level retrieval (e.g., Mem0~\cite{chhikara2025mem0}) often suffers from low information density, crowding out the finite context window with raw text rather than synthesized understanding.
Addressing these limitations, Generative Latent Memory has emerged as a promising paradigm~\cite{zhang2025memgen, xu2025softcot}.
By compressing broad contexts into dense continuous vectors, these methods function as compact cognitive buffers, preserving critical information without exhausting attention budgets.

However, a structural inefficiency persists in state-of-the-art latent memory paradigms due to \textit{architectural segregation} and \textit{computational redundancy}.
Prevalent methods typically synthesize memory via parameter-divergent auxiliary modules—whether independent encoders~\cite{wu2025towards, yu2025vismem} or specialized adapters~\cite{zhang2025memgen}—that are detached from the primary reasoning backbone.
This bifurcation prevents the direct reuse of the backbone's cached Key-Value (KV) states, compelling the system to maintain disjoint memory buffers and perform duplicate forward passes for context re-encoding.
Furthermore, relying on iterative optimization~\cite{li2025seek} or full-scale autoregressive decoding for memory generation inevitably interrupts the continuous flow of thought, introducing substantial inference latency.
We argue that an efficient memory mechanism should instead be \textit{intrinsic}—harvesting memory directly from the agent's existing computational flow to bypass redundant re-parameterization.

Realizing this vision, we propose \textbf{FlashMem}, a framework that distills latent memory directly from the agent's transient reasoning states via computation reuse.
Motivated by prior analyses suggesting that LLM final-layer representations retain rich trajectory information~\cite{nikolaou2025language}, FlashMem uses the last hidden state as a compact summary of the interaction history and as the seed for memory synthesis.
This allows us to employ a \textbf{Shared-KV Consolidator} that synthesizes memory by attending directly to the backbone's frozen cache, effectively treating the reasoning process itself as the definitive source of memory without structural segregation.
Complementing this, to align memory generation with real-time epistemic needs, we introduce a parameter-free \textbf{Cognitive Monitor}. Grounded in the correlation between attention entropy and model uncertainty~\cite{kuhn2023semantic, farquhar2024detecting}, this mechanism triggers consolidation exclusively when the agent exhibits high internal confusion.

Our contributions are summarized as follows:
\begin{itemize}
    \item We propose \textbf{FlashMem}, a lightweight framework designed to distill intrinsic memory via \textit{computation reuse}. Unlike segregated architectures that require heavy re-encoding, FlashMem directly harvests the backbone's transient reasoning states, offering a high-efficiency solution for long-horizon agents without architectural decoupling.

    \item We develop a minimalist \textbf{Shared-KV Consolidator} motivated by the observation that the last hidden state provides a compact summary for memory synthesis. By synthesizing memory directly from the backbone's frozen cache, this design eliminates redundancy. Empirical results demonstrate that it achieves performance parity with heavy baselines while reducing inference latency by \textbf{5$\times$} with negligible VRAM overhead.

    \item We introduce a parameter-free \textbf{Cognitive Monitor} that utilizes attention entropy as a real-time signal for epistemic uncertainty. Our analysis demonstrates that high entropy effectively localizes model confusion and hallucinations, enabling the monitor to adaptively trigger memory consolidation only when necessary to prevent wasteful computation.
\end{itemize}

\section{Related Work}
\label{sec:related_work}

Recent scholarship classifies agent memory into \textit{token-level}, \textit{parametric}, and \textit{latent} forms~\cite{hu2025memory}. While each has advanced the field, structural inefficiencies persist in how they are generated and managed.

\paragraph{Token-level and Parametric Memory.}
Token-level memory (e.g., RAG~\cite{lewis2020retrieval}, MemGPT~\cite{packer2023memgpt}, Mem0~\cite{chhikara2025mem0}) offers transparency but suffers from low information density, often exhausting finite attention budgets with retrieved context. Conversely, parametric memory (e.g., ROME~\cite{meng2022rome}) provides inference efficiency by internalizing knowledge into weights but lacks the plasticity required for real-time, intra-episode adaptation~\cite{hu2025memory}.

\paragraph{Generative Latent Memory.}
To improve density, research has shifted towards active memory generation, encoding context into continuous vectors.
Prominent works such as MemGen~\cite{zhang2025memgen} and SoftCoT~\cite{xu2025softcot} employ auxiliary modules—independent encoders or distinct LoRA adapters—to compress context into gist tokens~\cite{mu2023gist} or latent states.
Multimodal extensions like VisMem~\cite{yu2025vismem} and CoMEM~\cite{wu2025towards} similarly rely on separate encoders.
However, these architectures are computationally heavy.
For instance, MemGen necessitates maintaining three distinct model instances (Trigger, Weaver, and Reasoner) concurrently, while SoftCoT relies on autoregressive decoding or iterative optimization to synthesize prompts.
Consequently, they necessitate re-encoding history through secondary parameters, causing significant redundancy and VRAM overhead.
MemOS~\cite{li2025memos_long} is also related in its reuse of cached activations, but it focuses on reusable memory management rather than online synthesis of interaction-specific memory.

\paragraph{Efficiency and Cognitive Control.}
To mitigate the high costs of long-context reasoning, research focuses on two optimization paths: state compression and selective activation.
Heuristic methods like SnapKV~\cite{li2024snapkv}, H2O~\cite{zhang2023h2o}, and PyramidKV~\cite{cai2025pyramidkv} compress the KV cache by pruning less attentive tokens.
While fast, these extractive approaches risk severing long-range semantic dependencies compared to generative synthesis.
Beyond compression, optimizing activation timing is equally critical. Existing systems often rely on static prompts or heavy trainable classifiers~\cite{zhang2025memgen}, ignoring the established link between attention entropy and model uncertainty~\cite{kuhn2023semantic, farquhar2024detecting, guerreiro2023looking}.
FlashMem addresses these limitations by integrating a Shared-KV consolidator for efficient computation reuse and a parameter-free entropy monitor for self-regulated activation.

\begin{figure*}[t]
    \centering
    \includegraphics[width=\linewidth]{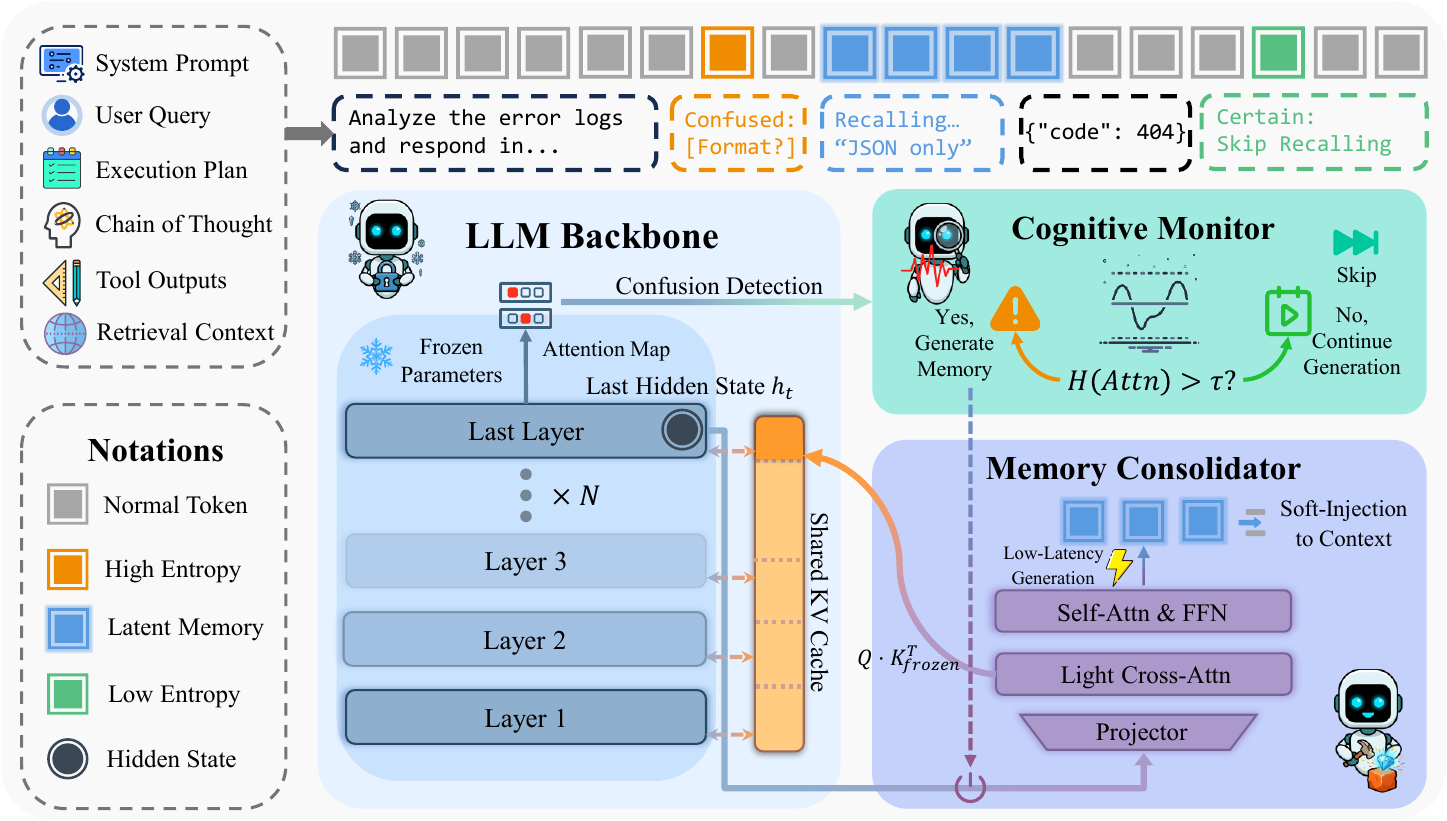}
    \caption{The architecture of \textbf{FlashMem}. It features a parameter-free \textbf{Cognitive Monitor} (top right) that activates memory generation based on attention entropy uncertainty, and a lightweight \textbf{Memory Consolidator} (bottom right). The consolidator utilizes a \textit{Shared-KV} design to efficiently generate latent memories by reusing the frozen backbone's cache, taking the hidden state $h_t$ as the initiation seed.}
    \label{fig:overview}
\end{figure*}

\section{Methodology}
\label{sec:method}

\subsection{Preliminaries and Problem Formulation}
\label{sec:preliminaries}

\paragraph{Agent Formulation and the Efficiency Gap.}
We formalize an LLM-based agent as a parametric policy $\pi_{\theta}$ interacting via a trajectory $\tau_{<t} = (o_1, a_1, \dots, o_{t-1}, a_{t-1})$. At step $t$, the agent samples action $a_t$ conditioned on the history $\tau_{<t}$, current observation $o_t$, and auxiliary memory $\mathcal{M}$:
\begin{equation}
    a_t \sim \pi_{\theta}(\cdot \mid \tau_{<t}, o_t, \mathcal{M})
\end{equation}
In this work, we focus on \textit{Latent Memory}, represented as continuous embeddings $\mathcal{M} = \{m_1, \dots, m_K\}$ ($m_i \in \mathbb{R}^d$).
Unlike token-level retrieval, latent memory functions as a high-density buffer that encodes abstract reasoning patterns with minimal token overhead.

However, prevalent paradigms typically adopt a segregated architecture, generating memory via an auxiliary process $\mathcal{G}_\phi$:
\begin{equation}
    \mathcal{M} = \mathcal{G}_\phi(\tau_{<t})
\end{equation}
This necessitates an independent encoding pass, incurring \textit{computational redundancy} via the repetitive encoding of $\tau_{<t}$ and \textit{state segregation} due to the maintenance of disjoint KV caches. Our goal is to synthesize $\mathcal{M}$ intrinsically by reusing the backbone's states.

\subsection{Overview of FlashMem}
\label{sec:overview}

To address the aforementioned efficiency gaps, we introduce \textbf{FlashMem}, a memory generation architecture that is deeply coupled with the LLM backbone $\pi_{\theta}$.
As illustrated in Figure~\ref{fig:overview}, unlike segregated systems that rely on independent encoding passes, FlashMem operates on the principle of computation reuse: it synthesizes memory directly from the backbone's transient activation states, thereby eliminating repetitive encoding and the overhead of maintaining disjoint KV caches.
The framework comprises two synergistic components:

\begin{itemize}
    \item \textbf{The Cognitive Monitor (Sec.~\ref{sec:monitor}):} A parameter-free gating mechanism that determines \textit{when} to generate memory. Upon detecting high entropy (signaling model confusion), the monitor suspends standard generation to initiate memory consolidation.

    \item \textbf{The Memory Consolidator (Sec.~\ref{sec:consolidator}):} A lightweight decoder designed to determine \textit{how} to synthesize memory efficiently. To avoid re-processing the long historical trajectory, it adopts a shared-KV design. Specifically, it projects the backbone's current hidden state $h_t$ to serve as the initial query, and then attends directly to the backbone's accumulated KV cache. This mechanism enables FlashMem to distill valuable knowledge and experiential patterns directly from the backbone's high-level representations, bypassing the need for redundant re-encoding.
\end{itemize}

\paragraph{Inference Workflow.}
During inference, FlashMem dynamically switches between standard generation and memory-augmented modes. The synthesized memory $\mathcal{M}$ is soft-injected back into the backbone's input stream. These latent embeddings serve as high-density cognitive cues, not only surfacing relevant historical context but also injecting abstract reasoning patterns to steer subsequent complex decision-making.

\subsection{The Cognitive Monitor: Entropy-Based Uncertainty Perception}
\label{sec:monitor}

Current generative memory paradigms often rely on rigid triggering mechanisms: either static injection---exemplified by SoftCoT~\cite{xu2025softcot} which indiscriminately prepends tokens---wasting computation on simple queries, or learned classifiers like MemGen~\cite{zhang2025memgen}, which introduce architectural overhead and distribution shift risks.
Instead, we propose to leverage the LLM's intrinsic uncertainty as a parameter-free activation signal.
Pioneering works in uncertainty estimation~\cite{kuhn2023semantic} and hallucination detection~\cite{farquhar2024detecting} have established that the dispersion of a model's internal probability distribution correlates strongly with its semantic confidence.
Specifically, when a model is confused or prone to confabulation, its attention mechanism tends to exhibit high entropy, lacking focus on salient tokens~\cite{guerreiro2023looking}.
Building on this theoretical consensus, we employ attention entropy as a real-time metacognitive indicator.

\paragraph{Mitigating Attention Sinks.}
However, calculating entropy directly from raw attention weights is unreliable due to the \textit{attention sink} phenomenon~\cite{xiao2024efficient}.
Studies show that LLMs (and Vision Transformers~\cite{darcet2024vision}) notoriously allocate massive attention scores to initial tokens (e.g., \texttt{[BOS]}), regardless of their semantic irrelevance. This dominant sink artificially acts as a low-entropy anchor, masking the true dispersion of the reasoning process.
To extract a de-noised cognitive signal, we process the attention weights from the last layer, which has been shown to encode the most semantic-rich and task-specific information~\cite{tenney2019bert}.
Let $A_{t, h} \in \mathbb{R}^{t}$ denote the attention distribution of head $h$ at the current step $t$ over the context sequence.
We define $\mathcal{S}_{\text{sink}}$ as the set of indices corresponding to the attention sinks. We explicitly mask these tokens and re-normalize the distribution for each head:
\begin{equation}
    \tilde{A}_{t, h}[j] = \frac{A_{t, h}[j] \cdot \mathbb{I}(j \notin \mathcal{S}_{\text{sink}})}{\sum_{k} A_{t, h}[k] \cdot \mathbb{I}(k \notin \mathcal{S}_{\text{sink}})}
\end{equation}
where $\mathbb{I}$ is the indicator function and $j, k$ represent token indices.

\paragraph{Entropy Decision and Adaptive Intervention.}
We calculate the Shannon entropy for each head individually to capture diverse uncertainty patterns, and then aggregate them to form the final metric. The system uncertainty $\mathcal{H}_t$ is defined as the average entropy across all $H$ heads:
\begin{equation}
    \mathcal{H}_t = \frac{1}{H} \sum_{h=1}^{H} \left( - \sum_{j \notin \mathcal{S}_{\text{sink}}} \tilde{A}_{t, h}[j] \log \tilde{A}_{t, h}[j] \right)
\end{equation}
Memory consolidation is activated only when $\mathcal{H}_t > \tau$.
The threshold $\tau$ functions as a sensitivity knob governing the efficiency-performance trade-off.
A lower $\tau$ encourages the agent to seek memory aid more aggressively (prioritizing accuracy), while a higher $\tau$ restricts intervention to moments of severe confusion (prioritizing speed). This flexibility allows FlashMem to adapt to varying computational budgets at test time, offering a distinct advantage over fixed-cost classifiers.
In practice, rather than relying on a fixed heuristic value, we adopt a distribution-aware calibration strategy.
The optimal baseline for $\tau$ correlates with the model's intrinsic confidence and the task's open-endedness.
We discuss these entropy dynamics and detail our statistical calibration protocol in Appendix~\ref{sec:appendix_threshold}.

\subsection{The Memory Consolidator: Shared-KV Generation}
\label{sec:consolidator}

Upon activation, the system aims to synthesize memory efficiently. To achieve this, we introduce the \textbf{Memory Consolidator}, a lightweight decoder architecture designed strictly around the principle of computation reuse.

\paragraph{State Projection from Final-Layer Representations.}
Motivated by prior work showing that final-layer representations preserve substantial information about the input trajectory~\cite{nikolaou2025language}, we use the backbone's last hidden state $h_t$ as a compact semantic summary of the preceding context.
Instead of re-encoding inputs, we treat the backbone as a pre-computed encoder. We project $h_t$ into the consolidator's working dimension via a two-layer MLP to produce the initial latent embedding $m_0$.
This allows the Consolidator to inherit the backbone's comprehension without redundant processing.

\paragraph{Projection-Free Cross-Attention.}
The Consolidator consists of $L$ stacked decoder layers. To minimize parameter count and VRAM usage, we introduce a projection-free cross-attention mechanism.
Unlike standard attention layers that require trainable projections for Keys and Values, we discard $W_K$ and $W_V$.
Let $x$ denote the input state to the current layer, initialized as $m_0$. The module computes only the Query vector using $x$ and attends directly to the backbone's accumulated key and value matrices $\mathbf{K}$ and $\mathbf{V}$:
\begin{equation}
    \text{Attn}(x, \mathbf{K}, \mathbf{V}) = \text{softmax}\left(\frac{(x W_Q) \mathbf{K}^\top}{\sqrt{d}}\right) \mathbf{V}
\end{equation}
where $W_Q$ is the only learnable projection matrix in the attention operation.
This mechanism ensures negligible storage overhead for context maintenance, as the generation process is anchored to the backbone's existing state pool.

\paragraph{Weight Inheritance and Soft Injection.}
Training a lightweight module from scratch to collaborate with a frozen, massive backbone presents a significant semantic alignment challenge. Random initialization often results in a large distributional shift, leading to unstable optimization.
To mitigate this, we employ a homologous weight inheritance strategy. Specifically, we initialize the Consolidator's parameters using the last $L$ layers of the backbone, where $L \ll N_{backbone}$.
This initialization places the Consolidator within a semantic manifold compatible with the backbone from the start, significantly accelerating convergence and facilitating smoother representation alignment.
The Consolidator then auto-regressively generates $K$ latent embeddings $\mathcal{M} = \{m_1, \dots, m_K\}$, which are soft-injected into the backbone's input stream to guide subsequent reasoning.

\section{Experiments}
\label{sec:experiments}

\subsection{Experimental Setup}
\label{sec:setup}
We evaluate FlashMem on six benchmarks: reasoning (GSM8K~\cite{cobbe2021training}, MATH~\cite{hendrycks2021measuring}, GPQA~\cite{rein2023gpqa}), coding (KodCode~\cite{xu2025kodcode}), and summarization (BookSum~\cite{kryscinski2022booksum}, GovReport~\cite{huang-etal-2021-efficient}).
We compare against three paradigms: (1) direct generation (Vanilla, CoT-SC~\cite{wang2023selfconsistency}); (2) KV compression (SnapKV~\cite{li2024snapkv}); and (3) generative latent memory (MemGen~\cite{zhang2025memgen}).
Experiments utilize instruct-tuned backbones from Qwen (2.5-1.5B, 3-4B)~\cite{qwen2.5,qwen3} and Llama (3.1-8B, 3.2-3B)~\cite{llama3.1,llama3.2}. A lightweight Consolidator is trained via SFT for each backbone while keeping the backbone frozen; detailed protocols and hyperparameters are in Appendix~\ref{sec:appendix_details}.
Appendix~\ref{sec:appendix_extra_results} also reports results on Qwen3-14B/32B, additional baselines (SoftCoT, LatentSeek, and MemOS), LongBench v2~\cite{bai2024longbench2} and RULER, and matched SFT controls.

\subsection{Main Results}
\label{sec:main_results}

Tables~\ref{tab:qwen_results} and \ref{tab:llama_results} present the performance of FlashMem against baselines on the Qwen and Llama families, respectively.

\begin{table*}[t]
\centering
\setlength{\aboverulesep}{0pt}
\setlength{\belowrulesep}{0pt}
\renewcommand{\arraystretch}{1.2}

\caption{
\textbf{Results on Qwen Family.} We compare FlashMem with baselines on Qwen 2.5 (1.5B) and the latest Qwen 3 (4B).
Metric: Accuracy (\%) for Reasoning/Code, ROUGE-1 for Summarization.
FlashMem consistently matches or exceeds the segregated latent memory baseline (MemGen) while maintaining superior efficiency.
Best results are \textbf{bolded}, and second-best results are \underline{underlined}.
}
\label{tab:qwen_results}
\resizebox{\textwidth}{!}{
\begin{tabular}{l|c|ccccc|ccccc}
\toprule
\multirow{2}{*}{\textbf{Dataset}} & \multirow{2}{*}{\textbf{Metric}} &
\multicolumn{5}{c|}{\cellcolor{qwenbg}\qwenlogo\textbf{Qwen 2.5 1.5B Instruct}} &
\multicolumn{5}{c}{\cellcolor{qwenbg}\qwenlogo\textbf{Qwen 3 4B Instruct}} \\
\cmidrule(lr){3-7} \cmidrule(lr){8-12}
 &  & \textbf{Vanilla} & \textbf{CoT-SC} & \textbf{SnapKV} & \textbf{MemGen} & \textbf{FlashMem} & \textbf{Vanilla} & \textbf{CoT-SC} & \textbf{SnapKV} & \textbf{MemGen} & \textbf{FlashMem} \\

\specialrule{\lightrulewidth}{3pt}{0pt}
\rowcolor{cattlerow} \multicolumn{12}{l}{\textbf{\textit{Mathematical \& Scientific Reasoning}}} \\
\arrayrulecolor{catline} \specialrule{0.5pt}{0pt}{2pt} \arrayrulecolor{black}
\textbf{GSM8K} & ACC & 65.12 & 68.98 & 54.37 & \textbf{70.54} & \underline{70.09} & 80.49 & 82.65 & 69.88 & \textbf{85.42} & \underline{84.38} \\
\textbf{MATH}  & ACC & 40.71 & 43.77 & 38.00 & \underline{46.55} & \textbf{50.16} & 54.22 & 54.96 & 50.12 & \textbf{60.23} & \underline{58.36} \\
\textbf{GPQA}  & ACC & 13.96 & \underline{15.14} & 13.66 & \textbf{16.13} & 14.73 & 16.84 & 17.95 & 14.54 & \textbf{21.68} & \underline{20.54} \\

\specialrule{\lightrulewidth}{3pt}{0pt}
\rowcolor{cattlerow} \multicolumn{12}{l}{\textbf{\textit{Code Generation}}} \\
\arrayrulecolor{catline} \specialrule{0.5pt}{0pt}{2pt} \arrayrulecolor{black}
\textbf{KodCode}& ACC & 22.85 & 29.30 & 39.70 & \textbf{55.70} & \underline{50.20} & 56.48 & 57.55 & 49.70 & \underline{59.77} & \textbf{61.13} \\

\specialrule{\lightrulewidth}{3pt}{0pt}
\rowcolor{cattlerow} \multicolumn{12}{l}{\textbf{\textit{Long-Context Summarization}}} \\
\arrayrulecolor{catline} \specialrule{0.5pt}{0pt}{2pt} \arrayrulecolor{black}

\textbf{BookSum} & R-1 & 11.14 & 11.62 & 10.38 & \underline{12.86} & \textbf{13.77} & 8.93 & 9.35 & \textbf{11.12} & 10.18 & \underline{10.99} \\
\textbf{GovReport}& R-1 & 13.61 & \underline{14.41} & 14.22 & 13.97 & \textbf{16.11} & 13.65 & 13.87 & 13.74 & \underline{14.31} & \textbf{15.52} \\
\bottomrule
\end{tabular}
}
\end{table*}

\begin{table*}[t]
\centering
\setlength{\aboverulesep}{0pt}
\setlength{\belowrulesep}{0pt}
\renewcommand{\arraystretch}{1.2}

\caption{
\textbf{Results on Llama Family.} Performance comparison on Llama 3.1 (8B) and Llama 3.2 (3B).
FlashMem demonstrates robust generalization across different parameter scales, particularly excelling in summarization and code tasks on smaller architectures.
Best results are \textbf{bolded}, and second-best results are \underline{underlined}.
}
\label{tab:llama_results}
\resizebox{\textwidth}{!}{
\begin{tabular}{l|c|ccccc|ccccc}
\toprule
\multirow{2}{*}{\textbf{Dataset}} & \multirow{2}{*}{\textbf{Metric}} &
\multicolumn{5}{c|}{\cellcolor{llamabg}\llamalogo\textbf{Llama 3.1 8B Instruct}} &
\multicolumn{5}{c}{\cellcolor{llamabg}\llamalogo\textbf{Llama 3.2 3B Instruct}} \\
\cmidrule(lr){3-7} \cmidrule(lr){8-12}
 &  & \textbf{Vanilla} & \textbf{CoT-SC} & \textbf{SnapKV} & \textbf{MemGen} & \textbf{FlashMem} & \textbf{Vanilla} & \textbf{CoT-SC} & \textbf{SnapKV} & \textbf{MemGen} & \textbf{FlashMem} \\

\specialrule{\lightrulewidth}{3pt}{0pt}
\rowcolor{cattlerow} \multicolumn{12}{l}{\textbf{\textit{Mathematical \& Scientific Reasoning}}} \\
\arrayrulecolor{catline} \specialrule{0.5pt}{0pt}{2pt} \arrayrulecolor{black}

\textbf{GSM8K} & ACC & 68.38 & 68.91 & 57.30 & \textbf{71.12} & \underline{70.46} & 60.48 & 60.76 & 55.62 & \textbf{63.68} & \underline{62.35} \\
\textbf{MATH}  & ACC & 45.97 & 50.37 & 44.23 & \underline{50.95} & \textbf{51.56} & 40.77 & 44.12 & 37.14 & \underline{45.17} & \textbf{48.05} \\
\textbf{GPQA}  & ACC & 15.48 & 17.80 & 15.66 & \textbf{18.43} & \underline{17.86} & 4.79 & 5.18 & 5.27 & \textbf{7.07} & \underline{6.70} \\
\specialrule{\lightrulewidth}{3pt}{0pt}
\rowcolor{cattlerow} \multicolumn{12}{l}{\textbf{\textit{Code Generation}}} \\
\arrayrulecolor{catline} \specialrule{0.5pt}{0pt}{2pt} \arrayrulecolor{black}
\textbf{KodCode}& ACC & 34.61 & 37.90 & 40.88 & \textbf{57.70} & \underline{54.27} & 21.93 & 26.27 & 20.33 & \underline{29.64} & \textbf{29.98} \\

\specialrule{\lightrulewidth}{3pt}{0pt}
\rowcolor{cattlerow} \multicolumn{12}{l}{\textbf{\textit{Long-Context Summarization}}} \\
\arrayrulecolor{catline} \specialrule{0.5pt}{0pt}{2pt} \arrayrulecolor{black}
\textbf{BookSum} & R-1 & 13.12 & 14.62 & 11.67 & \textbf{15.61} & \underline{14.65} & 9.03 & 9.29 & 10.25 & \textbf{12.09} & \underline{11.04} \\
\textbf{GovReport}& R-1 & 13.95 & 15.41 & 12.81 & \underline{16.37} & \textbf{17.23} & 12.92 & 13.61 & \underline{14.43} & 13.44 & \textbf{14.55} \\
\bottomrule
\end{tabular}
}
\end{table*}

\paragraph{Competitive Reasoning Fidelity.}
FlashMem achieves parity with the heavy-weight MemGen baseline across reasoning tasks without relying on independent encoders.
For instance, on Qwen 2.5 (Table~\ref{tab:qwen_results}), FlashMem reaches \textbf{70.09\%} on GSM8K, closely trailing MemGen (70.54\%) while significantly outperforming the Vanilla baseline.
Notably, on the smaller Llama 3.2 3B model (Table~\ref{tab:llama_results}), FlashMem even surpasses MemGen on the challenging MATH dataset (\textbf{48.05\%} vs. 45.17\%), suggesting that the shared-KV design may better preserve mathematical priors in smaller parametric spaces.

\paragraph{Long-Context Superiority and Robustness.}
FlashMem exhibits particular strength in summarization and structure-heavy tasks, consistently outperforming MemGen on GovReport and KodCode across most backbones.
On the Qwen 3 4B model, FlashMem leads on both KodCode (\textbf{61.13\%}) and GovReport (\textbf{15.52\%}), indicating that deriving memory directly from the backbone's high-fidelity KV cache captures global semantics better than re-encoding the history.
Furthermore, FlashMem significantly outperforms the KV compression baseline SnapKV (e.g., a \textbf{+11\%} gap on Qwen 3 KodCode), confirming that generative latent memory is essential for preserving the complex dependencies that heuristic pruning methods typically discard.

\paragraph{Broader Evaluation.}
This trend persists in the additional evaluations. On Qwen3-14B/32B, FlashMem remains close to MemGen on reasoning and improves on KodCode and GovReport, reaching 70.91 and 29.31, respectively, with Qwen3-32B. On Qwen3-4B, it also outperforms SoftCoT and LatentSeek on KodCode and GovReport while keeping much lower 64k latency. FlashMem further reaches 41.5 on LongBench v2 and 88.5 / 97.2 on RULER-64k NIAH / multi-hop tracing. Appendix~\ref{sec:appendix_extra_results} reports the full results.

\paragraph{MemOS and Matched SFT Controls.}
We also compare FlashMem with MemOS and matched SFT baselines. FlashMem is stronger than MemOS~\cite{li2025memos_long} on KodCode and GovReport, and is also more efficient in the compared 64k setting. Under the same expert trajectories, it outperforms both Full SFT and one-layer LoRA SFT on KodCode, suggesting that supervision alone does not account for the gain. Appendix~\ref{sec:appendix_extra_results} provides the detailed comparison.

\subsection{Efficiency Analysis}
\label{sec:efficiency}

\begin{table*}[t!]
\centering
\definecolor{overheadcolor}{HTML}{FFD6D6}
\caption{
\textbf{Efficiency Benchmark on Long-Context Inference.}
We report the \textbf{Peak VRAM} usage (maximum observed), along with \textbf{Effective Generation Throughput} (excluding latent tokens) and Average Latency across varying context lengths (4k to 64k).
Values are reported as Mean$_{\pm \text{Std. Dev.}}$.
All statistics are derived from 30 independent runs.
Background Color Intensity signifies performance overhead relative to the Vanilla baseline; darker red indicates higher overhead.
FlashMem maintains a profile nearly identical to the baseline, whereas MemGen incurs prohibitive costs at long contexts.
}
\label{tab:efficiency}
\resizebox{\textwidth}{!}{
\begin{tabular}{r|ccc|ccc|ccc}
\toprule
\multirow{2}{*}{\textbf{Context}} & \multicolumn{3}{c|}{\textbf{Peak Memory (GB)} $\downarrow$} & \multicolumn{3}{c|}{\textbf{Eff. Throughput (tokens/s)} $\uparrow$} & \multicolumn{3}{c}{\textbf{Latency (ms)} $\downarrow$} \\
\cmidrule(lr){2-4} \cmidrule(lr){5-7} \cmidrule(lr){8-10}
 & \textbf{Vanilla} & \textbf{MemGen} & \textbf{FlashMem} & \textbf{Vanilla} & \textbf{MemGen} & \textbf{FlashMem} & \textbf{Vanilla} & \textbf{MemGen} & \textbf{FlashMem} \\
\midrule
\textbf{4k}
& 4.61 & \cellcolor{overheadcolor!50} 13.19 & \cellcolor{overheadcolor!20} 7.29
& $39.45_{\pm 3.84}$ & \cellcolor{overheadcolor!50} $31.02_{\pm 2.87}$ & \cellcolor{overheadcolor!15} $37.14_{\pm 0.61}$
& $6.49_{\pm 0.94}$  & \cellcolor{overheadcolor!50} $8.25_{\pm 1.02}$  & \cellcolor{overheadcolor!15} $6.89_{\pm 0.10}$ \\
\textbf{8k}
& 8.66 & \cellcolor{overheadcolor!60} 15.03 & \cellcolor{overheadcolor!20} 8.90
& $38.29_{\pm 1.52}$ & \cellcolor{overheadcolor!60} $26.34_{\pm 1.93}$ & \cellcolor{overheadcolor!15} $37.28_{\pm 0.50}$
& $6.68_{\pm 0.45}$  & \cellcolor{overheadcolor!60} $9.72_{\pm 0.96}$  & \cellcolor{overheadcolor!15} $6.87_{\pm 0.09}$ \\
\textbf{16k}
& 11.88 & \cellcolor{overheadcolor!70} 18.71 & \cellcolor{overheadcolor!20} 12.12
& $37.34_{\pm 2.57}$ & \cellcolor{overheadcolor!75} $17.50_{\pm 1.16}$ & \cellcolor{overheadcolor!20} $35.65_{\pm 2.78}$
& $6.86_{\pm 0.82}$  & \cellcolor{overheadcolor!75} $14.63_{\pm 0.99}$ & \cellcolor{overheadcolor!20} $7.18_{\pm 0.68}$ \\
\textbf{32k}
& 18.32 & \cellcolor{overheadcolor!85} 26.07 & \cellcolor{overheadcolor!20} 18.56
& $34.24_{\pm 4.55}$ & \cellcolor{overheadcolor!90} $9.98_{\pm 0.24}$  & \cellcolor{overheadcolor!30} $29.74_{\pm 1.79}$
& $7.48_{\pm 1.33}$  & \cellcolor{overheadcolor!90} $25.65_{\pm 0.63}$ & \cellcolor{overheadcolor!30} $8.61_{\pm 0.51}$ \\
\textbf{64k}
& 31.21 & \cellcolor{overheadcolor!100} 40.78 & \cellcolor{overheadcolor!20} 31.44
& $25.67_{\pm 1.79}$ & \cellcolor{overheadcolor!100} $4.13_{\pm 0.06}$ & \cellcolor{overheadcolor!40} $20.86_{\pm 1.01}$
& $9.97_{\pm 0.83}$  & \cellcolor{overheadcolor!100} $61.99_{\pm 0.88}$& \cellcolor{overheadcolor!40} $12.28_{\pm 0.48}$ \\
\bottomrule
\end{tabular}
}
\end{table*}

To empirically validate the architectural efficiency of FlashMem, we conducted a stress test focusing on computational latency and memory footprint under long-context conditions.

\paragraph{Benchmark Protocol.}
We implemented a \textit{cyclic generation benchmark} on a single NVIDIA A100-80GB GPU.
The evaluation follows an iterative pattern of 8 cycles where the model generates 32 text tokens followed by 8 latent memory tokens.
We scaled the input context length $L$ from 4,096 (4k) to 65,536 (64k) tokens across Vanilla, MemGen~\cite{zhang2025memgen}, and FlashMem configurations.

\paragraph{Efficiency Results.}
As shown in Table~\ref{tab:efficiency}, the segregated architecture of MemGen exhibits poor scalability: at 64k tokens, its VRAM usage rises to \textbf{40.78 GB} (nearly 10 GB overhead), while effective throughput collapses to \textbf{4.13 tokens/s} due to re-encoding bottlenecks.
In contrast, FlashMem maintains a highly efficient profile nearly identical to the Vanilla baseline (31.44 GB vs 31.21 GB VRAM) while sustaining robust throughput (20.86 tokens/s).
Crucially, in terms of latency per step at maximum context, FlashMem (\textbf{12.28 ms}) achieves an approximate $\mathbf{5\times}$ speedup over MemGen (61.99 ms), confirming that the shared-KV design effectively eliminates redundant storage and computation.

\subsection{Impact of Consolidator Depth}
\label{sec:ablation}

To determine the optimal configuration for the Memory Consolidator, we investigated the effect of layer depth $L$ ($1 \le L \le 6$) on both reasoning performance and computational cost.

\begin{table}[h!]
\centering
\caption{
\textbf{Accuracy across Consolidator Depths.}
We report accuracy (\%) on GSM8K and KodCode for layer depths $L=1$ to $6$. Performance remains stable, indicating that increasing depth provides minimal benefit.
}
\label{tab:layer_acc}
\resizebox{0.95\linewidth}{!}{
\begin{tabular}{l|cccccc}
\toprule
Task & L=1 & L=2 & L=3 & L=4 & L=5 & L=6 \\
\midrule
GSM8K & 69.55 & 69.47 & 69.47 & 69.55 & 69.55 & 69.55 \\
KodCode & 50.10 & 50.19 & 50.29 & 50.19 & 50.29 & 50.29 \\
\bottomrule
\end{tabular}
}
\end{table}

\paragraph{Performance Saturation.}
Table~\ref{tab:layer_acc} details the accuracy on GSM8K and KodCode. We observe immediate performance saturation: the single-layer model ($L=1$) achieves 69.55\% on GSM8K, which is comparable to deeper configurations. Similarly, on the structure-sensitive KodCode task, increasing depth from $L=1$ to $L=6$ yields negligible gains (e.g., 50.10\% vs. 50.29\%).
This empirical evidence strongly suggests that a single-layer Consolidator is sufficient to distill high-quality latent memory from the backbone's shared KV cache, validating our minimalist design philosophy.

\begin{figure}[t!]
    \centering
    \includegraphics[width=\linewidth]{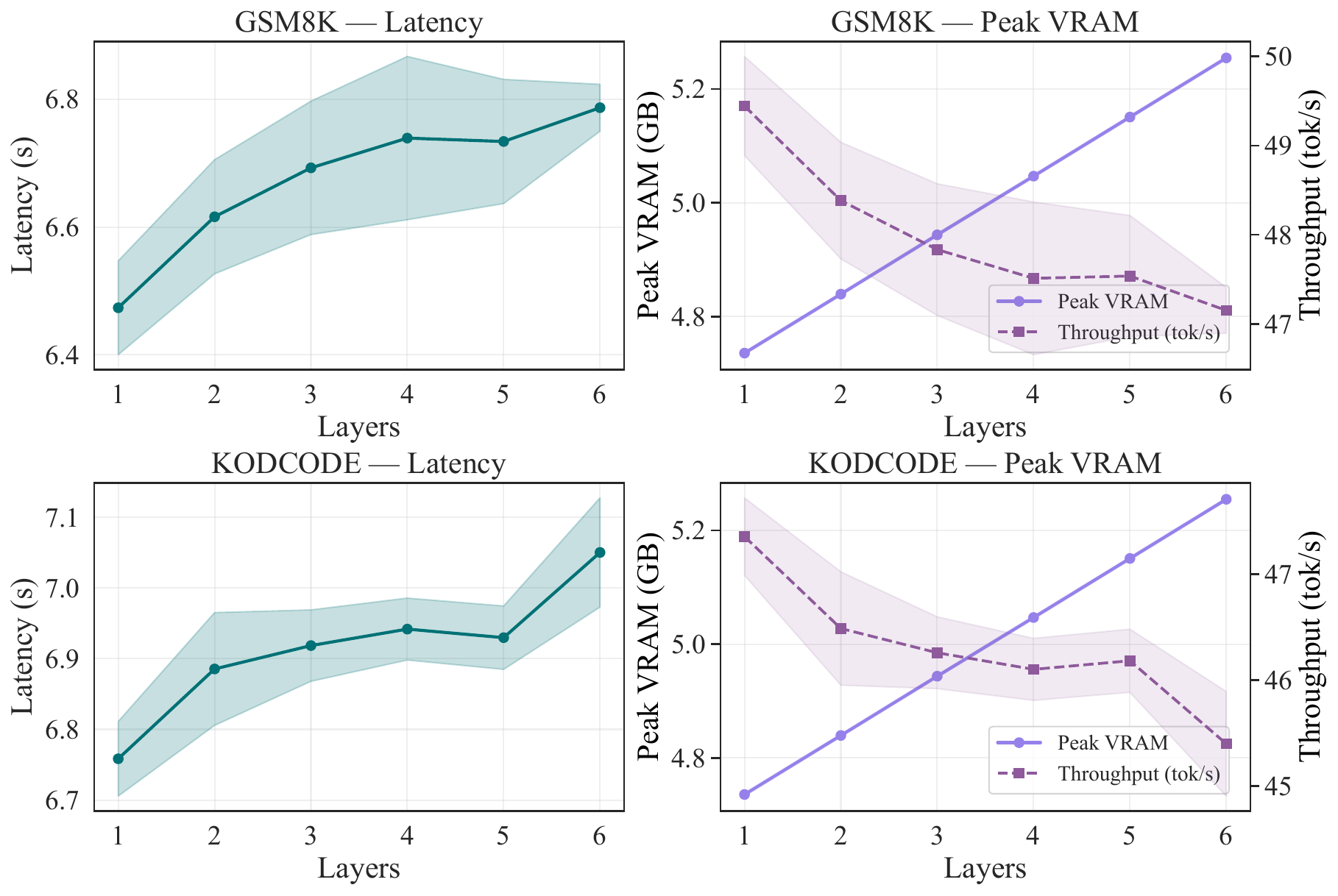}
    \caption{
    \textbf{Efficiency Trade-offs across Layer Depths.}
    VRAM usage scales linearly with depth, while latency increases monotonically. Combined with Table~\ref{tab:layer_acc}, results favor a minimal depth configuration.
    }
    \label{fig:layer_metrics}
\end{figure}

\paragraph{Cost Analysis.}
While accuracy plateaus, the computational overhead increases with depth. As illustrated in Figure~\ref{fig:layer_metrics}, VRAM usage exhibits a strict linear growth, directly correlating with the increase in parameters. In contrast, inference latency shows a monotonic increase (and throughput a corresponding decrease), reflecting the accumulated overhead of sequential decoding.
Given that $L=1$ offers nearly identical reasoning support with the lowest resource footprint, we conclude that shallow architectures are optimal for the FlashMem framework.

\section{Mechanism Analysis}
\label{sec:mechanism}

\subsection{Effectiveness of Cognitive Monitor}
\label{sec:monitor_analysis}

We hypothesize that the \textit{Cognitive Monitor} acts as an uncertainty detector, identifying critical decision points where the model lacks internal knowledge. To validate this, we analyzed inference traces on the GSM8K dataset, specifically focusing on samples where the Vanilla model failed while FlashMem succeeded.

\begin{figure}[h!]
    \centering
    \includegraphics[width=\linewidth]{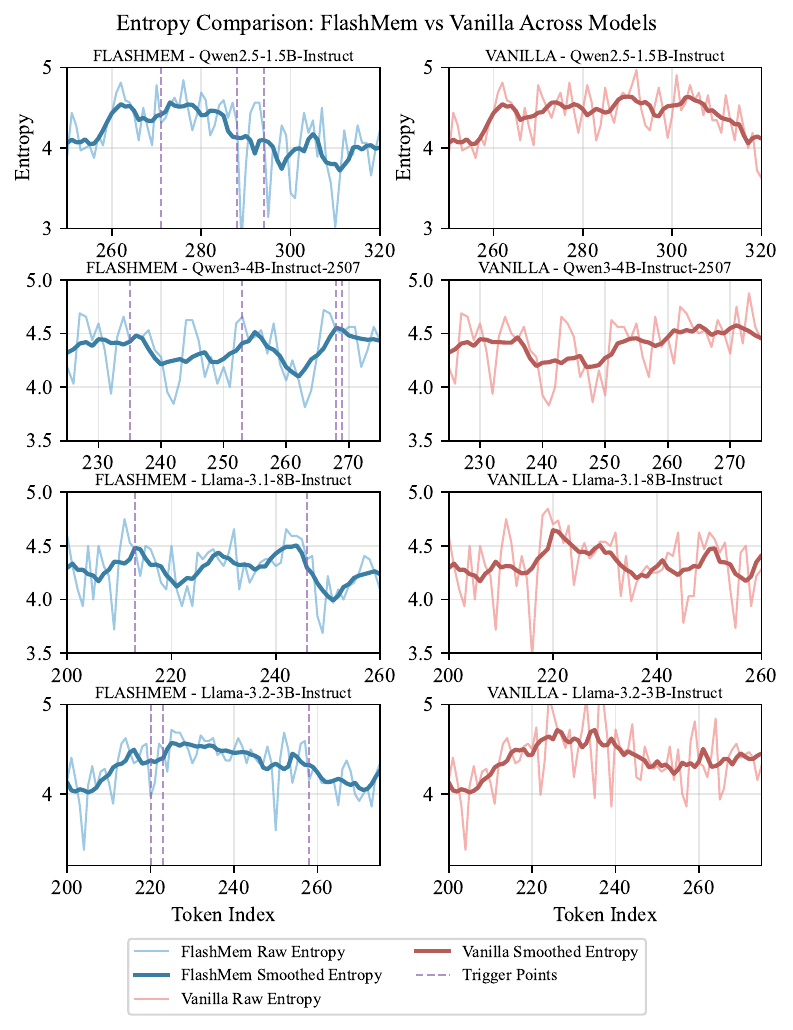}
    \caption{
    \textbf{Entropy Dynamics during Reasoning.}
    We visualize the attention entropy of FlashMem (Blue) vs. Vanilla (Red) across four models on challenging GSM8K samples.
    The vertical dashed lines indicate memory injection points triggered by the Cognitive Monitor. Post-injection, FlashMem exhibits a notable reduction in entropy compared to the baseline, reflecting increased confidence and reasoning stability.
    }
    \label{fig:entropy_vis}
\end{figure}

\paragraph{Qualitative Analysis: Entropy Dynamics.}
Figure~\ref{fig:entropy_vis} illustrates the attention entropy trajectories across four different backbones.
In these cases, the Vanilla baseline exhibits sustained or rising entropy at critical reasoning steps, indicative of cognitive confusion that eventually leads to logical errors.
In contrast, FlashMem successfully identifies these high-entropy peaks. Upon injecting the synthesized latent memory, we observe a pronounced reduction in attention entropy, reflecting a significant decrease in the model's internal uncertainty in subsequent tokens. This confirms that the Cognitive Monitor effectively pinpoints moments of confusion, and the inserted memory serves to stabilize the reasoning path. To rigorously validate this alignment, we further employed an LLM-as-a-Judge approach to localize the exact onset of reasoning errors, confirming that they highly coincide with the detected entropy peaks (see Appendix~\ref{sec:appendix_monitor_methods}).

\paragraph{Quantitative Analysis: Statistical Significance.}
To verify this entropy reduction is statistically robust, we computed metrics across all triggered instances in the GSM8K test set.

\begin{table}[h!]
\centering
\caption{
\textbf{Statistics of Entropy Reduction via Memory Injection.}
Metrics are calculated based on triggered steps in GSM8K. The high probability of reduction (76.5\%) confirms the consistent effectiveness of the Cognitive Monitor in mitigating model uncertainty.
}
\label{tab:entropy_stats}
\resizebox{\linewidth}{!}{
\begin{tabular}{l|c}
\toprule
\textbf{Metric} & \textbf{Statistics} \\
\midrule
Mean Entropy Reduction & 0.215 $\pm$ 0.125 \\
Mean Relative Reduction & 5.12\% $\pm$ 3.85\% \\
Entropy Reduction Range & -0.085 $\sim$ 0.850 \\
Probability of Reduction & 76.5\% \\
Prob. of Significant Reduction ($>0.5$) & 7.2\% \\
\bottomrule
\end{tabular}
}
\end{table}

The results are summarized in Table~\ref{tab:entropy_stats}.
On average, memory injection leads to a Mean Entropy Reduction of 0.215, with a relative reduction of 5.12\%. Notably, the Probability of Reduction is as high as 76.5\%, indicating that in the vast majority of cases, accessing latent memory effectively resolves the model's internal conflict. Furthermore, significant entropy drops (reduction $>0.5$) occur in 7.2\% of cases, which typically correspond to pivotal moments of insight where synthesized memory corrects a potential hallucination.

\subsection{Visualizing the Memory Consolidation Process}
\label{sec:viz_pointer}

To further investigate how FlashMem bridges the gap between distant context and current generation, we visualized the internal attention dynamics of both the Consolidator and the backbone.
Due to space constraints, the detailed qualitative analysis and the corresponding attention heatmaps are presented in Appendix~\ref{sec:appendix_vis}.
Our observations therein confirm that the Consolidator functions as an effective semantic filter, while the backbone maintains sustained attention on the injected latent tokens throughout the generation process.

\section{Conclusion}
\label{sec:conclusion}

In this paper, we introduced \textbf{FlashMem}, a framework that reconciles the tension between high-density latent memory and inference efficiency in autonomous agents. By replacing segregated auxiliary encoders with a \textbf{Shared-KV Consolidator}, FlashMem distills memory directly from the backbone’s frozen representations via computation reuse, eliminating the overhead of redundant re-parameterization. Furthermore, the integration of a parameter-free \textbf{Cognitive Monitor} empowers the agent to dynamically trigger memory consolidation based on real-time uncertainty, ensuring judicious resource allocation. Extensive experiments demonstrate that FlashMem matches the reasoning fidelity of heavy-weight baselines while reducing inference latency by up to $5\times$. We hope this work establishes a new paradigm for intrinsic, efficient memory mechanisms, facilitating the deployment of long-horizon agents in resource-constrained environments.

\section*{Limitations}

While FlashMem demonstrates robust efficiency and reasoning improvements, there are avenues for future research to further expand its scope.
First, our current framework is optimized for textual and code-based tasks.
We have not yet extended the Shared-KV paradigm to multimodal settings.
Given the increasing prominence of Vision-Language Models (VLMs), adapting FlashMem to handle multi-sensory inputs and visual tokens represents a promising direction for future work.
Second, while we have validated the effectiveness of FlashMem on representative open-source models, the exploration of this intrinsic memory mechanism on ultra-large-scale foundation models (e.g., those with hundreds of billions of parameters) remains to be conducted.
Investigating the scaling properties of computation reuse at such magnitudes would be valuable to fully map the potential of latent memory.

\section*{Ethical considerations}

This work aligns with Green AI principles by significantly lowering energy consumption through efficient computation reuse and reduced inference latency. We strictly adhere to the licenses and usage policies of the open-source models and datasets utilized in our experiments. While inheriting the inherent properties of pre-trained backbones, our framework does not introduce additional risks regarding data privacy or human rights violations.

\section*{Acknowledgments}
This work is partly funded by VUNI.2526.CAIR.10 and VUNI.GREEN-X.RISE.AY25-27.01.

\nocite{
    bian2026realmem,
    wang2023selfconsistency,
    kojima2022largelanguagemodels,
    wei2022chainofthought,
    geva2021transformer,
    dai2019transformerxl,
    kadavath2022language,
    lin2022teaching,
    liu2023scissorhands,
    ge2023model,
    borgeaud2022retro,
    wu2022memorizing,
    zhou2025onerecv2}
\bibliography{custom}

\clearpage
\appendix

\section{Experimental Implementation Details}
\label{sec:appendix_details}

\subsection{Data Construction and Training Objective}
\label{sec:appendix_training}

\paragraph{Domain-Adaptive Expert Distillation.}
To ensure the Memory Consolidator captures high-fidelity domain-specific reasoning patterns, we moved beyond generic pre-training corpora. We adopted a targeted expert distillation strategy. For each downstream domain—specifically \textit{Mathematical Reasoning} (GSM8K, MATH) and \textit{Code Generation} (KodCode)—we curated a dataset of high-quality execution trajectories.
To achieve this, we utilized advanced proprietary models to synthesize Chain-of-Thought (CoT) rationales. These synthesized trajectories were subjected to a rigorous rejection sampling filter, retaining only those that led to the correct final answer as verified by symbolic execution for code or deterministic equivalence checks for mathematics. This filtering process ensures that the Consolidator is trained exclusively on positive reinforcement signals, minimizing the risk of learning from hallucinated or suboptimal reasoning paths.

\paragraph{Sequence Construction and masking.}
For the optimization phase, we construct a unified training sequence $S$ for each instance. This sequence concatenates the initial instruction prompt $x$, the synthesized latent memory placeholders $\mathcal{M}_{gen}$, and the ground-truth expert trajectory $y$:
\begin{equation}
    S = [x, \mathcal{M}_{gen}, y]
\end{equation}
Here, $\mathcal{M}_{gen}$ represents the sequence of $K$ latent vectors generated by the Consolidator. Crucially, we apply a strategic masking protocol to the loss computation. The label indices corresponding to the instruction tokens $x$ and the latent memory tokens $\mathcal{M}_{gen}$ are set to \texttt{-100}, which serves as the standard ignore index in PyTorch. Consequently, the loss is calculated only on the prediction of the expert trajectory $y$.

\paragraph{Gradient Flow and Optimization.}
The training objective is to minimize the standard Cross-Entropy Loss ($\mathcal{L}_{CE}$) on the target tokens $y$, conditioned on the injected memory:
\begin{equation}
    \mathcal{L} = - \sum_{t=1}^{|y|} \log P_{\theta}(y_t \mid x, \mathcal{M}_{gen}, y_{<t})
\end{equation}
A critical aspect of our implementation is the computation graph. The backbone parameters $\theta$ remain entirely frozen ($\nabla_\theta = 0$). Gradients are back-propagated from the loss function, through the backbone's frozen layers, and exclusively into the soft memory embeddings $\mathcal{M}_{gen}$. From there, they flow into the Consolidator's parameters $\psi$. This setup forces the Consolidator to learn an implicit objective: it must encode information into $\mathcal{M}_{gen}$ such that the frozen backbone's likelihood of generating the correct expert trajectory $y$ is maximized. This effectively functions as knowledge distillation from the future, compressing the semantic essence of the trajectory $y$ into the latent state $\mathcal{M}_{gen}$.

\subsection{Hyperparameters and Hardware Environment}
\label{sec:appendix_hyperparams}

\paragraph{Infrastructure and Software Stack.}
All training and inference experiments were conducted on a high-performance computational node equipped with 4 $\times$ NVIDIA A100-SXM4-80GB GPUs. The framework was implemented using PyTorch 2.7.1 and the Hugging Face Transformers library. To optimize training throughput and manage memory footprints for the larger backbone models, we utilized DeepSpeed ZeRO-2 optimization and FlashAttention-2 kernels. This allowed us to train the Consolidator efficiently without the need for full model sharding in smaller scale experiments.

\paragraph{Configuration Justification.}
Table~\ref{tab:hyperparams} provides a comprehensive listing of the hyperparameters.
We selected a memory token count of $K=8$ as an optimal operating point for our architecture.
Regarding the Consolidator depth, based on the efficiency benchmarks detailed in Section~\ref{sec:ablation}, we set $L=1$ to prioritize minimal VRAM overhead and maximum throughput.
To mitigate the risk of exploding gradients—a known instability in autoregressive memory generation—we enforced a strict gradient clipping norm of 0.53. The learning rate schedule utilized a linear warmup for the first 10\% of steps followed by a cosine decay, ensuring the stable adaptation of the parameters inherited from the backbone within the pre-trained semantic space.

\begin{table}[h!]
\centering
\caption{Hyperparameters for FlashMem Training and Inference.}
\label{tab:hyperparams}
\resizebox{0.95\linewidth}{!}{
\begin{tabular}{l|c}
\toprule
\textbf{Hyperparameter} & \textbf{Value} \\
\midrule
\multicolumn{2}{l}{\textit{Model Architecture}} \\
Memory Token Count ($K$) & 8 \\
Consolidator Layers ($L$) & 1 \\
Hidden Dimension & Matches Backbone \\
\midrule
\multicolumn{2}{l}{\textit{Training Optimization}} \\
Optimizer & AdamW \\
Learning Rate & 1e-5 \\
Weight Decay & 0.01 \\
Gradient Clipping & 0.53 \\
Global Batch Size & 64 \\
Training Epochs & 5 \\
Scheduler & Cosine Decay \\
Warmup Ratio & 0.1 \\
\midrule
\multicolumn{2}{l}{\textit{Inference}} \\
Memory Decoding Strategy & Greedy Search (Deterministic) \\
Entropy Threshold ($\tau$) & Adaptive (85th percentile) \\
\bottomrule
\end{tabular}
}
\end{table}

\subsection{Inference Dynamics and Algorithm}
\label{sec:appendix_inference}

The runtime behavior of FlashMem is governed by an interactive loop that alternates between standard token generation and latent memory consolidation. We provide the formal pseudocode in Algorithm~\ref{alg:inference}.

\paragraph{Deterministic Memory Generation.}
Unlike the standard stochastic sampling used for text generation, the memory generation step (Line 8 in Algorithm~\ref{alg:inference}) employs a strict Greedy Search strategy. This deterministic approach serves two purposes: first, it stabilizes the latent space, preventing the injection of high-variance noise into the backbone's context; second, it ensures reproducibility of the reasoning trajectory given the same context state.

\paragraph{Cache Management and Injection.}
A critical implementation detail lies in the management of the Key-Value (KV) Cache (Line 10). When memory $\mathcal{M}$ is synthesized, it is not merely concatenated to the input prompt string. Instead, the continuous embeddings of $\mathcal{M}$ are fed through the backbone to generate their corresponding KV pairs, which are then appended to the active KV cache $\mathcal{K}, \mathcal{V}$. This allows the backbone to attend to these virtual memory tokens in all subsequent steps as if they were part of the physical context window, without requiring re-computation of the prior history.

\begin{algorithm}[h!]
\small
\caption{FlashMem Inference Loop}
\label{alg:inference}
\begin{algorithmic}[1]
\State \textbf{Input:} Prompt $x$, Backbone $\pi_\theta$, Consolidator $\mathcal{C}_\psi$, Threshold $\tau$
\State \textbf{Initialize:} KV Cache $\mathcal{K}, \mathcal{V} \gets \emptyset$, Context $c \gets x$
\While{not end of generation}
    \State Logits, Attentions, $h_{last} \gets \pi_\theta(c, \mathcal{K}, \mathcal{V})$
    \State $\mathcal{H}_{current} \gets \text{CalcEntropy}(\text{Attentions})$

    \If{$\mathcal{H}_{current} > \tau$} \Comment{High Uncertainty Detected}
        \State $m_0 \gets \text{Project}(h_{last})$
        \State $\mathcal{M} \gets \text{GreedyGen}(\mathcal{C}_\psi, m_0, \mathcal{K}, \mathcal{V}, \text{steps}=8)$
        \State \textbf{Inject:} Soft-append $\mathcal{M}$ to Backbone Input
        \State Update $\mathcal{K}, \mathcal{V}$ with $\mathcal{M}$ states
    \EndIf

    \State $x_{next} \gets \text{Sample}(\text{Logits})$
    \State $c \gets x_{next}$
\EndWhile
\end{algorithmic}
\end{algorithm}

\section{Additional Experimental Results}
\label{sec:appendix_extra_results}

This section reports the additional experiments referenced in the main paper, including larger backbones and expanded baselines, retrieval-heavy long-context evaluation, comparison with MemOS, and attribution analysis with matched SFT baselines. Tables~\ref{tab:appendix_qwen_large}--\ref{tab:appendix_sft_attr} summarize the corresponding results.

\subsection{Scaling to Larger Backbones and Broader Baselines}
\label{subsec:appendix_scaling_expanded}

We first examine whether the main-paper results hold on larger Qwen3 backbones and under stronger latent-memory comparisons. Table~\ref{tab:appendix_qwen_large} shows that FlashMem remains close to MemGen on GSM8K while improving on KodCode and GovReport. Table~\ref{tab:appendix_expanded_baselines} compares FlashMem with SoftCoT~\cite{xu2025softcot} and LatentSeek~\cite{li2025seek} on Qwen3-4B.

\begin{table*}[t]
\centering
\small
\caption{Results on larger Qwen3 backbones.}
\label{tab:appendix_qwen_large}
\resizebox{\textwidth}{!}{
\begin{tabular}{llccc}
\toprule
\textbf{Model} & \textbf{Method} & \textbf{GSM8K Acc. $\uparrow$} & \textbf{KodCode Acc. $\uparrow$} & \textbf{GovReport R-1 $\uparrow$} \\
\midrule
Qwen3-14B & Vanilla  & 86.64 & 63.17 & 19.43 \\
Qwen3-14B & CoT-SC   & 88.53 & 65.76 & 22.67 \\
Qwen3-14B & SnapKV   & 84.59 & 60.35 & 18.71 \\
Qwen3-14B & MemGen   & 92.24 & 67.81 & 24.28 \\
Qwen3-14B & FlashMem & 92.18 & 68.72 & 24.57 \\
\midrule
Qwen3-32B & Vanilla  & 89.71 & 66.93 & 23.14 \\
Qwen3-32B & CoT-SC   & 91.46 & 68.04 & 24.48 \\
Qwen3-32B & SnapKV   & 87.68 & 63.21 & 22.93 \\
Qwen3-32B & MemGen   & 94.11 & 69.44 & 26.44 \\
Qwen3-32B & FlashMem & 93.75 & 70.91 & 29.31 \\
\bottomrule
\end{tabular}
}
\end{table*}

\begin{table*}[t]
\centering
\small
\caption{Expanded latent-memory baseline comparison on Qwen3-4B.}
\label{tab:appendix_expanded_baselines}
\resizebox{\textwidth}{!}{
\begin{tabular}{lccccc}
\toprule
\textbf{Method} & \textbf{GSM8K Acc. $\uparrow$} & \textbf{KodCode Acc. $\uparrow$} & \textbf{GovReport R-1 $\uparrow$} & \textbf{Latency@64k (ms) $\downarrow$} & \textbf{VRAM@64k (GB) $\downarrow$} \\
\midrule
Vanilla    & 80.49 & 56.48 & 13.65 & $9.97_{\pm 0.83}$  & 31.21 \\
MemGen     & 85.42 & 59.77 & 14.31 & $61.99_{\pm 0.88}$ & 40.78 \\
SoftCoT    & 82.15 & 53.62 & 13.92 & $48.50_{\pm 2.10}$ & 33.50 \\
LatentSeek & 84.10 & 59.44 & 15.05 & $55.26_{\pm 3.15}$ & 39.80 \\
FlashMem   & 84.38 & 61.13 & 15.52 & $12.28_{\pm 0.48}$ & 31.44 \\
\bottomrule
\end{tabular}
}
\end{table*}

Tables~\ref{tab:appendix_qwen_large} and \ref{tab:appendix_expanded_baselines} show the same trade-off as in the main paper: FlashMem stays close to MemGen on reasoning, improves on KodCode and GovReport, and keeps much lower long-context overhead than SoftCoT and LatentSeek.

\subsection{Retrieval-Heavy Long-Context Evaluation}
\label{subsec:appendix_longbench_ruler}

To complement BookSum and GovReport, we additionally evaluate FlashMem on LongBench v2 and RULER-64k~\cite{hsieh2024ruler}. Table~\ref{tab:appendix_longbench_ruler} reports the results on these retrieval- and tracing-oriented long-context benchmarks.

\begin{table}[t]
\centering
\small
\caption{Retrieval-heavy long-context evaluation on Qwen3-32B.}
\label{tab:appendix_longbench_ruler}
\resizebox{\linewidth}{!}{
\begin{tabular}{llcccc}
\toprule
\textbf{Benchmark} & \textbf{Metric} & \textbf{Vanilla} & \textbf{SnapKV} & \textbf{MemGen} & \textbf{FlashMem} \\
\midrule
LongBench v2 & Overall Accuracy $\uparrow$ & 38.6 & 38.2 & 40.1 & 41.5 \\
RULER (NIAH) & Accuracy (\%) $\uparrow$ & 86.2 & 81.5 & 87.8 & 88.5 \\
RULER (Multi-hop Tracing) & Accuracy (\%) $\uparrow$ & 96.0 & 92.3 & 96.7 & 97.2 \\
\bottomrule
\end{tabular}
}
\end{table}

As shown in Table~\ref{tab:appendix_longbench_ruler}, FlashMem performs best on all three benchmarks, suggesting that its memory remains useful beyond summarization settings.

\subsection{Comparison with MemOS}
\label{subsec:appendix_memos}

We next compare FlashMem with MemOS~\cite{li2025memos_long}, which is related through cache-based memory reuse. Table~\ref{tab:appendix_memos_perf} reports task performance, and Table~\ref{tab:appendix_memos_eff} reports 64k efficiency.

\begin{table*}[t]
\centering
\small
\caption{Task-level comparison with MemOS.}
\label{tab:appendix_memos_perf}
\resizebox{\textwidth}{!}{
\begin{tabular}{llccc}
\toprule
\textbf{Model} & \textbf{Method} & \textbf{GSM8K Acc. $\uparrow$} & \textbf{KodCode Acc. $\uparrow$} & \textbf{GovReport R-1 $\uparrow$} \\
\midrule
Qwen2.5-1.5B & Vanilla  & 65.12 & 22.85 & 13.61 \\
Qwen2.5-1.5B & SnapKV   & 54.37 & 39.70 & 14.22 \\
Qwen2.5-1.5B & MemOS    & 66.84 & 42.65 & 14.58 \\
Qwen2.5-1.5B & MemGen   & 70.54 & 55.70 & 13.97 \\
Qwen2.5-1.5B & FlashMem & 70.09 & 50.20 & 16.11 \\
\midrule
Qwen3-4B & Vanilla  & 80.49 & 56.48 & 13.65 \\
Qwen3-4B & SnapKV   & 69.88 & 49.70 & 13.74 \\
Qwen3-4B & MemOS    & 82.17 & 57.32 & 14.26 \\
Qwen3-4B & MemGen   & 85.42 & 59.77 & 14.31 \\
Qwen3-4B & FlashMem & 84.38 & 61.13 & 15.52 \\
\bottomrule
\end{tabular}
}
\end{table*}

\begin{table}[t]
\centering
\small
\caption{64k efficiency comparison with MemOS on Qwen2.5-1.5B.}
\label{tab:appendix_memos_eff}
\resizebox{\linewidth}{!}{
\begin{tabular}{lccccc}
\toprule
\textbf{Metric} & \textbf{Vanilla} & \textbf{SnapKV} & \textbf{MemOS} & \textbf{MemGen} & \textbf{FlashMem} \\
\midrule
Peak VRAM (GB) $\downarrow$ & 31.21 & 22.54 & 35.62 & 40.78 & 31.44 \\
Throughput (tok/s) $\uparrow$ & $25.67_{\pm 1.79}$ & $28.32_{\pm 1.56}$ & $18.74_{\pm 1.12}$ & $4.13_{\pm 0.06}$ & $20.86_{\pm 1.01}$ \\
Latency (ms) $\downarrow$ & $9.97_{\pm 0.83}$ & $8.92_{\pm 0.75}$ & $13.66_{\pm 0.94}$ & $61.99_{\pm 0.88}$ & $12.28_{\pm 0.48}$ \\
\bottomrule
\end{tabular}
}
\end{table}

Together, Tables~\ref{tab:appendix_memos_perf} and \ref{tab:appendix_memos_eff} show that FlashMem is more effective for dynamic interaction memory in our setting, while MemOS is better aligned with reusable memory management.

\subsection{Attribution Analysis with Matched SFT Baselines}
\label{subsec:appendix_sft_attr}

Finally, we ask whether FlashMem's gains can be explained by supervision on expert trajectories alone. Table~\ref{tab:appendix_sft_attr} compares FlashMem against controlled SFT baselines trained on the same trajectories, which separates the effect of supervision from the effect of the memory mechanism.

\begin{table}[t]
\centering
\small
\caption{Controlled attribution analysis on Qwen2.5-1.5B using identical expert trajectories.}
\label{tab:appendix_sft_attr}
\resizebox{\linewidth}{!}{
\begin{tabular}{lcccc}
\toprule
\textbf{Method} & \textbf{GSM8K Acc. $\uparrow$} & \textbf{KodCode Acc. $\uparrow$} & \textbf{Training Peak VRAM (GB) $\downarrow$} & \textbf{Inference Peak VRAM@64k (GB) $\downarrow$} \\
\midrule
Vanilla & 65.12 & 22.85 & -- & 31.21 \\
Full SFT & 67.83 & 38.62 & 28.64 & 31.21 \\
LoRA SFT (1-Layer) & 66.95 & 32.17 & 5.50 & 31.21 \\
MemGen & 70.54 & 55.70 & 13.50 & 40.78 \\
FlashMem (1-Layer) & 70.09 & 50.20 & 5.98 & 31.44 \\
\bottomrule
\end{tabular}
}
\end{table}

As shown in Table~\ref{tab:appendix_sft_attr}, both Full SFT and one-layer LoRA SFT remain substantially weaker than FlashMem on KodCode under identical expert trajectories. This suggests that supervision alone does not account for the observed gain.

\section{Methodology for Cognitive Monitor Analysis}
\label{sec:appendix_monitor_methods}

This section details the experimental protocols employed to validate the efficacy of the Cognitive Monitor. These protocols underpin the quantitative and qualitative analyses presented in Section~\ref{sec:monitor_analysis}, specifically focusing on the correlation between attention entropy and reasoning fidelity.

\subsection{Error Localization via LLM-as-a-Judge}
\label{subsec:llm_judge}

To verify that detected high-entropy peaks correspond to genuine cognitive breakpoints and internal model confusion, we conducted a fine-grained error analysis on the GSM8K dataset.

\paragraph{Automated Evaluation Setup.}
Manually annotating the precise onset of reasoning errors across thousands of trajectories is prohibitively expensive and prone to subjective bias. Consequently, we employed Gemini 3 Pro as an objective, automated evaluator (LLM-as-a-Judge). We instructed the judge to audit the reasoning trajectories generated by the unaugmented Vanilla model with the specific objective of pinpointing the exact reasoning step where the logic first deviated from the ground truth or exhibited internal inconsistency.

The prompt utilized for this evaluation was designed to enforce strict step-by-step verification, as shown in Table~\ref{tab:judge_prompt}.

\begin{table}[h!]
    \centering
    \small
    \renewcommand{\arraystretch}{1.3}
    \begin{tabular}{|p{0.95\linewidth}|}
        \hline
        \textbf{System Prompt:} You are an expert Mathematics Evaluator. Your task is to inspect a given solution trajectory for a math problem and identify the first step where a logical or calculation error occurs. \\
        \\
        \textbf{Input:}
        \begin{itemize}
            \setlength\itemsep{0em}
            \item \textbf{Question:} [Input Question]
            \item \textbf{Gold Solution:} [Ground Truth Reference]
            \item \textbf{Model Trajectory:} [Model's Reasoning Output]
        \end{itemize}
        \\
        \textbf{Instructions:}
        \begin{enumerate}
            \setlength\itemsep{0em}
            \item Compare the Model Trajectory against the Gold Solution step-by-step.
            \item Identify the first step index (0-indexed) where the reasoning is flawed. If the reasoning is strictly correct but diverges in method, do not count it as an error unless it leads to a wrong answer.
            \item If the entire trajectory is correct, return -1.
        \end{enumerate}
        \\
        \textbf{Output Format:} \\
        Return a JSON object: \texttt{\{"error\_found": boolean, "first\_error\_step": int, "reason": "string"\}} \\
        \hline
    \end{tabular}
    \caption{The evaluation prompt used for the LLM-as-a-Judge error localization analysis.}
    \label{tab:judge_prompt}
\end{table}

\paragraph{Temporal Alignment Analysis.}
We collected the error timestamps returned by the judge and aligned them with the time-series entropy logs recorded during the generation of the Vanilla trajectories. This alignment revealed a distinct synchronization pattern where significant spikes in attention entropy consistently precede or coincide with the steps identified as erroneous. This empirical evidence validates our core premise that attention entropy serves as a reliable real-time proxy for model confusion, providing the physical basis for the unstable high-entropy behavior observed in the Vanilla baseline (Figure~\ref{fig:entropy_vis}).

\subsection{Statistical Protocol for Entropy Reduction}
\label{subsec:entropy_stats}

Table~\ref{tab:entropy_stats} in the main text reports the statistical impact of FlashMem on reducing model uncertainty. Here, we define the paired control experiment designed to isolate the causal effect of memory injection on attention entropy.

\paragraph{Data Construction.}
To ensure a fair comparison, we utilized a twin-run data collection strategy. We collected inference traces from two distinct execution modes on the same set of evaluation prompts:
\begin{itemize}
    \item FlashMem Trajectories (Treatment Group): Data collected from the model equipped with the full FlashMem mechanism. These traces include an augmentation mask that explicitly logs the timestamp $t$ of every memory injection event triggered by the Cognitive Monitor.
    \item Vanilla Trajectories (Control Group): Data collected from the base model without memory augmentation. We enforced the model to process the same prompts to use the Vanilla generation as a counterfactual baseline for entropy levels without intervention.
\end{itemize}

\paragraph{Window-based Comparative Analysis.}
Entropy is a dynamic metric that fluctuates naturally with token distinctness. To systematically quantify the immediate local impact of memory injection and filter out global noise, we employed a window-based difference-in-differences statistical approach:

\begin{enumerate}
    \item Trigger Identification: For each sample in the FlashMem group, we identified the trigger index $t$ corresponding to the activation of the augmentation mask where the Cognitive Monitor determined $\mathcal{H} > \tau$. Valid triggers were restricted to indices $t > 5$ to mitigate the confounding effects of initial generation instability.

    \item Window Definition: We established a local analysis window $W$ with a fixed horizon of $L=10$ tokens subsequent to the trigger point, encompassing the interval $[t, t+L]$. This window captures the immediate downstream effect of the injected memory on the internal state of the backbone.

    \item Metric Calculation: We computed the mean attention entropy $\mathcal{H}$ within this specific window for the FlashMem trajectory and contrasted it with the mean entropy of the Vanilla baseline measured at identical positional indices. This point-to-point comparison neutralizes variance caused by position-dependent entropy shifts.
\end{enumerate}

\paragraph{Quantification Metrics.}
The net reduction effect for each triggered sample $i$ is quantified as the difference between the counterfactual baseline and the treated state:
\begin{equation}
    \Delta \mathcal{H}_i = \mathcal{H}_{\text{vanilla}}^{(i)}[t:t+L] - \mathcal{H}_{\text{flashmem}}^{(i)}[t:t+L]
\end{equation}
A positive $\Delta \mathcal{H}_i$ indicates that the injection of memory successfully lowered model uncertainty. Based on the distribution of $\Delta \mathcal{H}_i$, we derived the following statistical indicators:
\begin{itemize}
    \item Probability of Reduction: Defined as $P(\Delta \mathcal{H}_i > 0)$. This represents the frequency with which memory injection successfully stabilizes the attention distribution.
    \item Probability of Significant Reduction: The proportion of samples where $\Delta \mathcal{H}_i > \tau_{sig}$ (with $\tau_{sig}=0.5$). These instances typically correspond to critical moments where the synthesized memory resolves major ambiguity.
    \item Relative Reduction Percentage: A normalized metric calculated as $\frac{\Delta \mathcal{H}_i}{\mathcal{H}_{\text{vanilla}}^{(i)}} \times 100\%$, reflecting the magnitude of the improvement relative to the initial confusion level.
\end{itemize}
This paired design ensures that the observed entropy drop is causally linked to the FlashMem intervention, confirming that the latent context directly enhances model certainty rather than being an artifact of token positioning.

\section{Analysis of Entropy Dynamics and Threshold Selection}
\label{sec:appendix_threshold}

The Cognitive Monitor regulates the initiation of memory consolidation via a decision threshold $\tau$. Our empirical analysis indicates that absolute attention entropy is not a universal constant; rather, it fluctuates significantly depending on the underlying model architecture and the structural properties of the prompt. In this section, we delineate these governing dynamics and detail the statistical methodology employed for adaptive threshold determination.

\subsection{Empirical Trends in Attention Entropy}

\paragraph{Model Scale and Confidence}
We observed a distinct inverse correlation between model parameter count and intrinsic entropy levels. Larger language models typically manifest sharper attention distributions, concentrating probability mass on fewer tokens. This behavior reflects a higher degree of internal confidence in next-token predictions compared to smaller architectures, which often exhibit more diffuse attention patterns. Consequently, a static threshold suitable for a smaller model would be insufficiently sensitive for a larger one. To maintain consistent monitoring performance, the absolute value of $\tau$ must scale inversely with model size, ensuring that the system remains responsive to relative drops in confidence rather than absolute magnitude.

\paragraph{Task Topology and Solution Space}
The baseline entropy profile is further modulated by the topology of the task solution space. Tasks characterized by vast, open-ended search spaces, such as code generation or creative writing, naturally induce higher entropy fluctuations due to the validity of multiple divergent generation paths. In contrast, tasks requiring convergent reasoning, such as arithmetic or strict logic, typically operate within a lower entropy regime. Summarization and general knowledge retrieval generally fall between these extremes. These variations imply that the definition of a high-uncertainty state is relative. A specific entropy value indicating confusion in a math task might represent a normal operating state in a creative writing context.

\subsection{Distribution-Aware Selection Strategy}
In light of these architectural and task-dependent variations, relying on a fixed heuristic value for $\tau$ is suboptimal. We instead adopt a distribution-aware strategy to calibrate the activation boundary dynamically.

We conceptualize memory injection as an intervention mechanism specifically designed for cognitive outliers—instances where the model deviates significantly from its standard confidence baseline. To capture these outliers accurately, we calibrate $\tau$ based on the entropy distribution observed on a representative validation subset prior to inference. Formally, the threshold is determined via the percentile function:

\begin{equation}
    \tau = \text{Percentile}(\{\mathcal{H}_i\}_{i=1}^N, P_{\text{target}})
\end{equation}

In our experiments, we configure $P_{\text{target}}$ to target the upper tail of the distribution, specifically the 85th percentile. This configuration serves as an effective filter, suppressing routine generation noise while preserving sensitivity to significant uncertainty peaks. These peaks typically correspond to critical decision points where the model requires external memory support. This statistical approach allows FlashMem to automatically adapt the absolute threshold to the specific characteristics of the model and task distribution, eliminating the need for manual fine-tuning.

\section{Detailed Visualization of Attention Dynamics}
\label{sec:appendix_vis}

\begin{figure}[t!]%
    \centering
    \includegraphics[width=\linewidth]{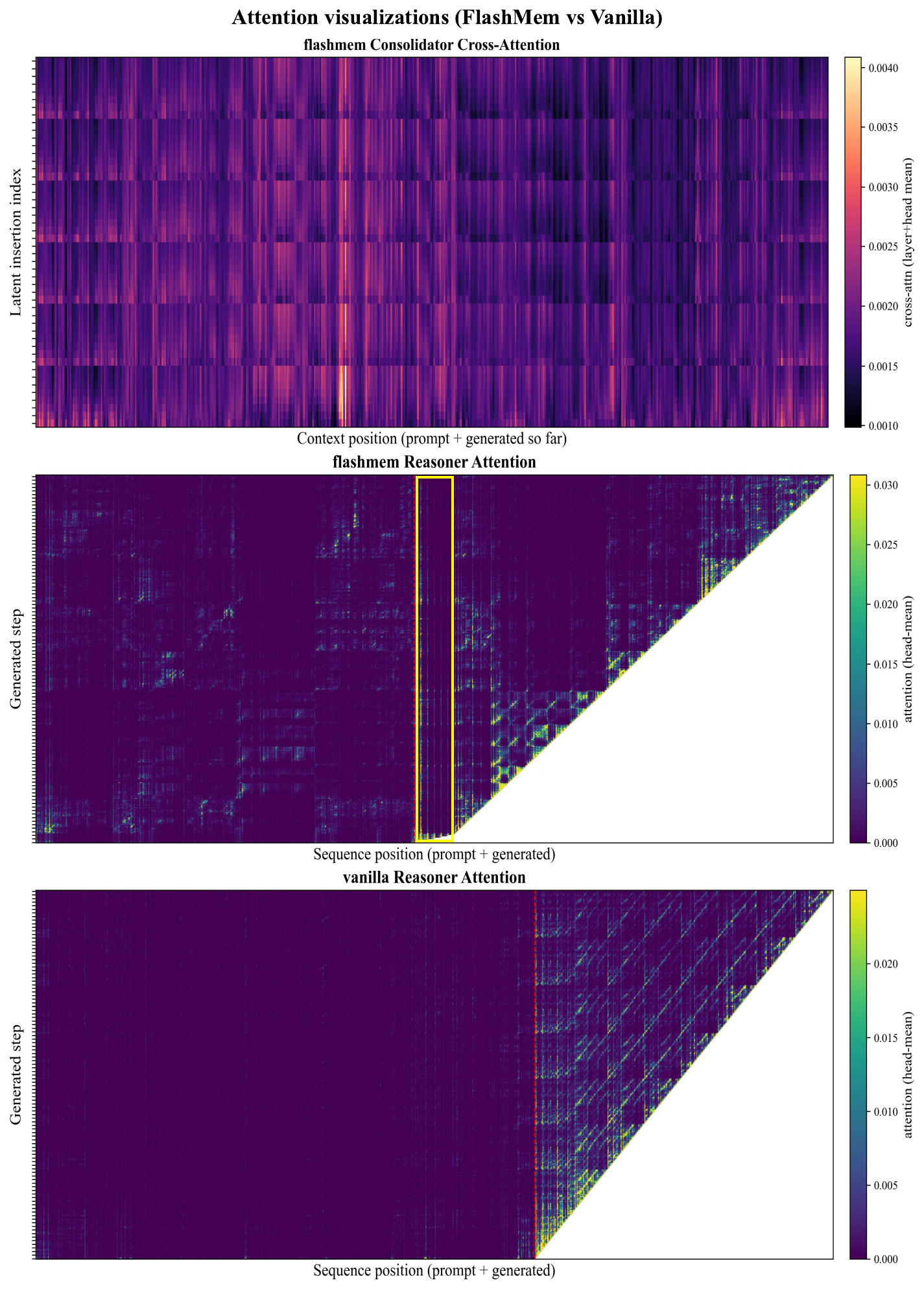}
    \caption{
    \textbf{Visualization of Attention Dynamics.}
    \textbf{Top:} Consolidator's cross-attention accurately targets key information (e.g., variable definitions) in the context.
    \textbf{Middle:} FlashMem's backbone consistently attends to the injected latent tokens (marked by yellow boxes) throughout generation, as evidenced by the continuous vertical attention lines.
    \textbf{Bottom:} The Vanilla model exhibits diffuse attention, failing to capture distant dependencies.
    }
    \label{fig:attention_map}
\end{figure}
In this section, we provide the detailed qualitative analysis and visualization regarding the KodCode error case mentioned in Section~\ref{sec:mechanism}.

\paragraph{Semantic Extraction via Cross-Attention.}
The Top Panel of Figure~\ref{fig:attention_map} illustrates the Cross-Attention weights of the Memory Consolidator during latent token generation.
We observe that the Consolidator exhibits highly selective attention, focusing intensely on specific regions of the input context (highlighted in the heatmap).
A detailed inspection reveals that these high-weight regions align with semantically salient segments within the extensive history, which are essential for the current prediction. This confirms that the Shared-KV mechanism effectively functions as a semantic filter, extracting and compressing key dependencies from the raw history into the latent space.

\paragraph{Sustained Reference via Self-Attention.}
The Middle Panel visualizes the Self-Attention of the FlashMem backbone during the subsequent text generation phase.
To facilitate observation, we inserted latent tokens at fixed intervals (marked by a yellow box).
The red dashed line demarcates the boundary between the input prompt and the model-generated text.%
A striking pattern emerges: there are five distinct vertical attention bands aligned perfectly with the inserted latent tokens.
This indicates that throughout the generation process, the backbone consistently attends back to these latent embeddings.
These memory tokens effectively function as high-density semantic anchors, providing continuous support for ongoing reasoning.

\paragraph{Comparison with Vanilla Baseline.}
In contrast, without the guidance of latent memory, the attention distribution appears significantly more diffuse and lacks structural focus.
The model fails to maintain focus on the distant variable definitions (which appear dark in the map), leading to the loss of critical context and the subsequent hallucination of undefined variables.
This comparison strongly validates our hypothesis: FlashMem succeeds not by simply extending context length, but by creating explicit, accessible semantic bridges that the model actively utilizes.

\section{Extended Qualitative Analysis}
\label{sec:appendix_cases}

In this section, we present a detailed analysis of failure modes in mathematical reasoning using the GSM8K benchmark. We select three representative cases from the test set where the baseline model fails due to input hallucination, variable binding errors, or over-reasoning.

In the comparative tables below, we explicitly identify the decision points where the FlashMem cognitive monitor triggers a latent memory injection. These interventions serve to refresh critical context and steer the reasoning trajectory back to the correct logical path compared to the baseline.

\subsection{Correction of Input Information Hallucination}

Reasoning tasks often require precise extraction of numerical values from natural language descriptions. A common failure mode is input hallucination, where the model correctly identifies the entity but assigns it an incorrect value derived from internal priors.

Table~\ref{tab:qualitative_comparison} (Case 1) illustrates a time calculation problem. The prompt specifies that peeling a potato takes "a minute and a half" (90 seconds). The baseline model hallucinates a value of "1 minute and 15 seconds." In the FlashMem trajectory, the memory mechanism triggers immediately at the unit conversion step. This injection acts as a semantic anchor, enforcing adherence to the specific text span in the input history, thereby overriding the hallucination.

\subsection{Rectification of Variable Binding}

Complex problems often involve multiple subsets of a population. Models frequently struggle with variable binding, applying a rate or ratio to the wrong subset after intermediate calculations.

Table~\ref{tab:qualitative_comparison} (Case 2) presents a party planning scenario. The constraint is explicit: gift bags are needed for each "invited guest." The baseline model calculates the attendees correctly but then erroneously applies the gift bag ratio to this new number. For FlashMem, the memory injection occurs after the attendee calculation. This intervention refreshes the original constraint regarding the invited guests, correcting the operand selection for the final cost estimation.

\subsection{Mitigation of Semantic Over-Reasoning}

Large language models can sometimes over-reason, interpreting straightforward problem statements as trick questions or applying irrelevant real-world constraints that contradict the mathematical abstraction.

Table~\ref{tab:qualitative_comparison} (Case 3) details a problem asking for egg consumption. The baseline model performs the arithmetic correctly but then rejects the result in the final step, arguing that "making" is not "eating." FlashMem performs the same calculations, but when approaching the conclusion, the memory injection stabilizes the model's focus on the quantitative accumulation task, suppressing the generation of the semantic refusal.

\begin{table*}[p]%
    \centering
    \footnotesize%
    \renewcommand{\arraystretch}{1.15}%
    \setlength{\tabcolsep}{4pt}

    \resizebox{\textwidth}{!}{
    \begin{tabular}{p{0.48\textwidth}|p{0.48\textwidth}}
    \toprule
    \multicolumn{2}{l}{\textsc{Case 1: Input Information Hallucination}} \\
    \multicolumn{2}{p{0.96\textwidth}}{Problem: Billy has 60 potatoes. It takes him \textbf{a minute and a half} to peel a potato, and 5 seconds to cut it. How long will it take to finish prepping the potatoes?} \\
    \midrule
    \textsc{Baseline Generation} & \textsc{FlashMem Generation} \\
    \midrule
    Let's break down the time per potato. & \textcolor{blue}{\textit{[Latent Memory Injection Triggered]}} \\
    First, convert the peeling time. & $\hookrightarrow$ \textit{Retrieving precise duration: "minute and a half" implies 90s.} \\
    1 minute and 15 seconds = 75 seconds. & Let's break down the time per potato. \\
    The time to cut is 5 seconds. & First, convert the peeling time. \\
    & 1 minute and 30 seconds = 60 + 30 = 90 seconds. \\
    So, total time per potato: & The time to cut is 5 seconds. \\
    $75 + 5 = 80$ seconds. & So, total time per potato: \\
    & $90 + 5 = 95$ seconds. \\
    Now for 60 potatoes: & Now for 60 potatoes: \\
    $60 \times 80 = 4800$ seconds. & $60 \times 95 = 5700$ seconds. \\
    Convert to minutes: $4800 / 60 = 80$. & Convert to minutes: $5700 / 60 = 95$. \\
    Final Answer: 80 & Final Answer: 95 \\
    \bottomrule
    \end{tabular}
    }

    \vspace{1em}%

    \resizebox{\textwidth}{!}{
    \begin{tabular}{p{0.48\textwidth}|p{0.48\textwidth}}
    \toprule
    \multicolumn{2}{l}{\textsc{Case 2: Variable Binding Error}} \\
    \multicolumn{2}{p{0.96\textwidth}}{Problem: Christina needs 0.75 gift bags per \textbf{invited guest}. She invited 16 friends, but 1/4 don't show up. Gift bags are \$2 each. How much will she spend?} \\
    \midrule
    \textsc{Baseline Generation} & \textsc{FlashMem Generation} \\
    \midrule
    Step 1: Determine guests attending. & Step 1: Determine guests attending. \\
    Total invited = 16. & Total invited = 16. \\
    Guests not showing up = $1/4 \times 16 = 4$. & Guests not showing up = $1/4 \times 16 = 4$. \\
    Guests attending = $16 - 4 = 12$. & Guests attending = $16 - 4 = 12$. \\
    \midrule
    Step 2: Determine gift bags needed. & \textcolor{blue}{\textit{[Latent Memory Injection Triggered]}} \\
    We need 0.75 bags per guest. & $\hookrightarrow$ \textit{Refreshing constraint: Ratio applies to "Invited" (16), not "Attending".} \\
    So, $0.75 \times 12 = 9$ bags. & Step 2: Determine gift bags needed. \\
    & The problem states 0.75 bags per \textbf{invited guest}. \\
    & So, $0.75 \times 16 = 12$ bags. \\
    \midrule
    Step 3: Calculate total cost. & Step 3: Calculate total cost. \\
    $9 \times \$2 = \$18$. & $12 \times \$2 = \$24$. \\
    Final Answer: 18 & Final Answer: 24 \\
    \bottomrule
    \end{tabular}
    }

    \vspace{1em}%

    \resizebox{\textwidth}{!}{
    \begin{tabular}{p{0.48\textwidth}|p{0.48\textwidth}}
    \toprule
    \multicolumn{2}{l}{\textsc{Case 3: Semantic Over-Reasoning}} \\
    \multicolumn{2}{p{0.96\textwidth}}{Problem: Claire makes a 3 egg omelet every morning. How many dozens of eggs will she eat in 4 weeks?} \\
    \midrule
    \textsc{Baseline Generation} & \textsc{FlashMem Generation} \\
    \midrule
    Step 1: Calculate daily usage. & Step 1: Calculate daily usage. \\
    She uses 3 eggs per day. & She uses 3 eggs per day. \\
    Step 2: Calculate 4 weeks usage. & Step 2: Calculate 4 weeks usage. \\
    4 weeks = 28 days. & 4 weeks = 28 days. \\
    Total eggs = $3 \times 28 = 84$ eggs. & Total eggs = $3 \times 28 = 84$ eggs. \\
    Step 3: Convert to dozens. & Step 3: Convert to dozens. \\
    $84 / 12 = 7$ dozens. & $84 / 12 = 7$ dozens. \\
    \midrule
    Step 4: Final Verification. & \textcolor{blue}{\textit{[Latent Memory Injection Triggered]}} \\
    Wait, the question asks what she will \textit{eat}. & $\hookrightarrow$ \textit{Suppressing distractor: "Making" implies "Eating" in this math context.} \\
    She is making them, not eating them. & Step 4: Final Verification. \\
    Therefore, the answer is 0. & The calculation holds. She consumes 7 dozens. \\
    Final Answer: 0 & Final Answer: 7 \\
    \bottomrule
    \end{tabular}
    }

    \caption{Qualitative comparison on GSM8K. We display the complete reasoning chain for both models. The blue markers indicate the exact moment the FlashMem mechanism intervenes. The accompanying text describes the semantic correction provided by the memory to the ongoing generation state.}
    \label{tab:qualitative_comparison}
\end{table*}

\end{document}